\documentclass[letterpaper, 10 pt, conference]{ieeeconf}  

\IEEEoverridecommandlockouts                              

\usepackage{amsmath} 
\usepackage{amssymb}  
\usepackage{pifont}
\usepackage{graphicx}
\usepackage{caption}
\usepackage{makecell}
\usepackage{booktabs}
\usepackage{threeparttable}
\usepackage{dirtree}
\usepackage{hyperref}
\usepackage{url}
\usepackage{algpseudocode}

\title{\LARGE \bf
SurgPose: a Dataset for Articulated Robotic Surgical Tool Pose Estimation and Tracking}

\author{Zijian Wu$^{1}$, Adam Schmidt$^{2}$, Randy Moore$^{1}$, Haoying Zhou$^{3}$, Alexandre Banks$^{1}$, \\Peter Kazanzides$^{4}$, and Septimiu E. Salcudean$^{1}$
\thanks{$^{1}$ Zijian Wu, Randy Moore, Alexandre Banks, and Septimiu E. Salcudean are from the Robotics and Control Laboratory (RCL), the University of British Columbia, Vancouver, Canada
        {\tt\small \{zijianwu, tims\}@ece.ubc.ca}}%
\thanks{$^{2}$ Adam Schmidt is from Intuitive Surgical,
        Sunnyvale, USA}
\thanks{$^{3}$ Haoying Zhou is from Worcester Polytechnic Institute, Worcester, USA}
\thanks{$^{4}$ Peter Kazanzides is from Johns Hopkins University,
        Baltimore, USA}
}

\let\oldtwocolumn\twocolumn
\renewcommand\twocolumn[1][]{%
    \oldtwocolumn[{#1}{
    \begin{center}
           \includegraphics[width=\textwidth]{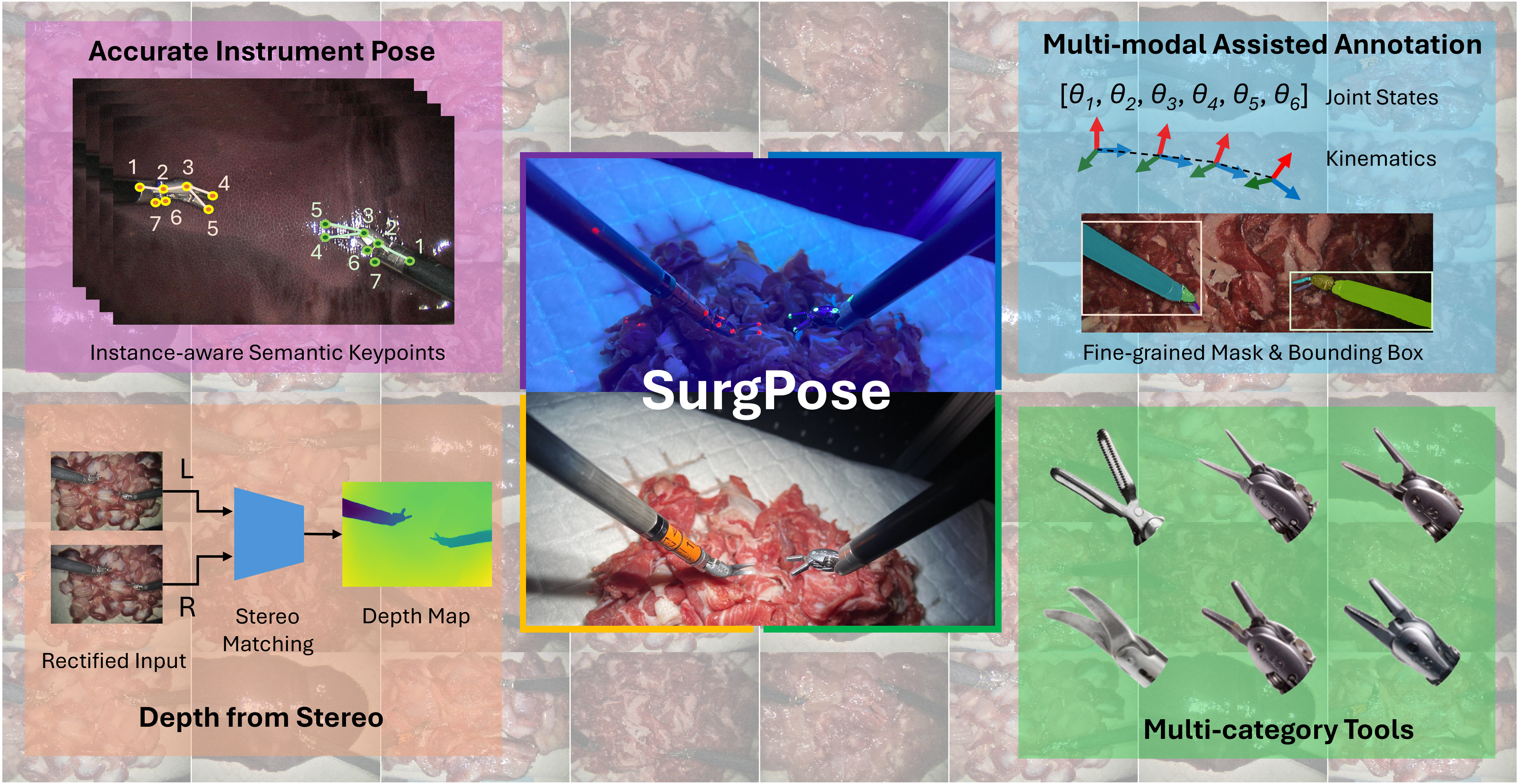}
           \captionof{figure}{Overview diagram of the proposed SurgPose dataset. We collect 30 trajectories with diverse annotations using a da Vinci IS1200 system with the da Vinci Research Kit (dVRK), with \textit{ex vivo} tissue background. We mark keypoints on 6 classes of surgical tools using ultraviolet (UV) reactive paint. Each trajectory is acquired under both UV and white light to extract keypoint annotations.}
           \label{fig1}
        \end{center}
    }]
}

\begin{document}

\maketitle
\thispagestyle{empty}
\pagestyle{empty}



\begin{abstract}

Accurate and efficient surgical robotic tool pose estimation is of fundamental significance to downstream applications such as augmented reality (AR) in surgical training and learning-based autonomous manipulation. 
While significant advancements have been made in pose estimation for humans and animals, it is still a challenge in surgical robotics due to the scarcity of published data. The relatively large absolute error of the da Vinci end effector kinematics and arduous calibration procedure make calibrated kinematics data collection expensive. 
Driven by this limitation, we collected a dataset, dubbed SurgPose, providing instance-aware semantic keypoints and skeletons for visual surgical tool pose estimation and tracking.
By marking keypoints using ultraviolet (UV) reactive paint, which is invisible under white light and fluorescent under UV light, we execute the same trajectory under different lighting conditions to collect raw videos and keypoint annotations, respectively.
The SurgPose dataset consists of approximately 120k surgical instrument instances ($\thicksim$80k for training and $\thicksim$40k for validation) of 6 categories.
Each instrument instance is labeled with 7 semantic keypoints. 
Since the videos are collected in stereo pairs, the 2D pose can be lifted to 3D based on stereo-matching depth. 
In addition to releasing the dataset, we test a few baseline approaches to surgical instrument tracking to demonstrate the utility of SurgPose.
More details can be found at \href{https://surgpose.github.io/}{surgpose.github.io}.
\end{abstract}


\section{INTRODUCTION}
Robot-assisted surgery (RAS) is revolutionizing traditional surgery. Surgical robotic systems incorporating advanced imaging technologies can reduce the surgeon's effort and improve outcomes compared with standard laparoscopic surgery, especially for complicated operations~\cite{kockerling2014robotic}. Intuitive Surgical's da Vinci Robotic System is the most prevalent product in RAS. 
Having instrument pose measurements for this system, and others, enables a wide spectrum of downstream applications including augmented reality (AR)~\cite{kalia2021preclinical}, multi-modal data registration~\cite{song2024arc,wu2023automatic}, visual servoing of the endoscopic camera~\cite{9217086}, and learning-based manipulation~\cite{kim2024surgical}. 
The joint states and kinematics of the da Vinci system robotic arms are accessible in those systems that have the da Vinci API. In real-world practice, however, the joint states and rigid kinematics models do not provide accurate instrument localization because of cable and instrument flexibility and backlash.
In addition, for instruments to appear at their actual location in the console camera view, a hand-eye calibration is necessary. 
Previous attempts to solve this hand-eye calibration problem include iterative methods~\cite{zhang2017computationally}, interactive methods~\cite{kalia2019marker, 8963603}, and recurrent neural networks~\cite{hwang2020efficiently}. 
However, these calibration procedures are onerous due to the long kinematic chains of the da Vinci Patient Side Manipulators (PSMs), which have seven degrees of freedom (DoF), 18 joints, and over two meters of cumulative length. 
Driven by these limitations, a camera-based method that directly computes the instrument's pose with respect to the laparoscopic camera becomes an attractive solution. 

Deep learning-based approaches have outperformed traditional methods in a series of scene understanding tasks, achieving state-of-the-art results~\cite{Hu_2023_CVPR, peng2023openscene}. 
Fully supervised learning depends on datasets with high-quality annotations to achieve good performance. 
The remarkable progress in deep learning would not have been possible without training on large-scale datasets. 
The recent success of the foundation model (FM)~\cite{kirillov2023segment} and the large language model (LLM)~\cite{zhao2023survey} are manifestations of the ``scaling law". However, robotics applications, particularly in surgical scenarios, face a shortage of high-quality training and validation data due to highly diverse data distributions and the expensive cost of data collection in the physical world. Generating synthetic data for training can facilitate this problem through sim-to-real transfer. However, the sim-to-real gap limits performance, and there remains a need for real-world datasets to ensure effective evaluation.

To overcome barriers to training and validation of deep learning models in robot-assisted surgery, we construct a dataset, SurgPose, for articulated surgical tool pose estimation. SurgPose has the following advantages:
\begin{itemize}
\item Transparent markers to provide precise keypoint position while causing little image difference between before and after painting. 
\item Multi-modal assisted annotations, including end-effector kinematics, joint states of the PSM, and part-level segmentation masks. These annotations also have the potential for other tasks, such as joint configuration estimation, kinematics recovery, etc. 
\item Rectified stereo, which enables lifting 2D poses to 3D.
\item 6 types of surgical instruments, which enables more comprehensive evaluation for algorithms.
\end{itemize}

In the following, we present related work (Section II), our methodology (Section III), experiments with publicly available pose estimation methods and suggestions for future use (Section IV), and, limitations of our dataset and future work (Section V). 

\begin{table*}[h]
\caption{Comparison of SurgPose with existing Robotic Surgical Instrument Pose Estimation
datasets.}
\label{table_1}
\begin{center}
\begin{tabular}{c c c c c c c c c c}
\toprule
\makecell[l]{Dataset\\\, } & \makecell[l]{Stereo\\\,} & \makecell[l]{Kinematics \&\\ Joint States} & \makecell[l]{Camera\\ Matrix} & \makecell[l]{Type\\\,} & \makecell[l]{\# Types of\\ Instruments} & \makecell[l]{Trajectories\\\,} & \makecell[l]{Labeled\\ Frames} & \makecell[l]{Labeled\\Instances} & \makecell[l]{Keypoints\\\,} \\
\midrule
\makecell[l]{EndoVis15~\cite{bodenstedt2018comparative}} & \ding{55} & \ding{55} & \ding{55} & \textit{ex vivo} & 2 & 6 & 9K & 12K & 1 \\
\makecell[l]{ Du et al., 2018~\cite{du2018articulated}} & \ding{55} & \ding{55} & \ding{55} & \textit{ex vivo} & 2 & 6 & 1.85K & 24.52K & 5 \\
\makecell[l]{SurgPose (Ours)} & \ding{51} & \ding{51} & \ding{51} & \textit{ex vivo} & 6 & 30 & 60K & 120K & 7\\
\bottomrule
\end{tabular}
\end{center}
\end{table*}

\section{Related Work}

\subsection{Surgical Tool Pose Estimation and Tracking}
SurgPose is designed for developing and evaluating surgical tool pose estimation and tracking. Template matching is a commonly used method for articulated surgical tool pose estimation and tracking. Low-level visual features are extracted to learn templates by machine learning algorithms~\cite{reiter2014appearance}. Reiter et al.~\cite{reiter2012articulated} proposed a method using the CAD model and kinematics of the tool to generate the virtual template. Ye et al.~\cite{ye2016real} achieved real-time tracking by introducing an online virtual tool renderer. 
Allan et al.~\cite{allan20183} used silhouette and optical flow-based features to estimate the 3D pose of the robotic surgical instrument.
However, these low-level feature representations suffer from a lack of semantic information, which causes performance degradation when facing corrupted images and different surgical scenes.
Deep learning-based methods have demonstrated better robustness and generalizability in surgical scene understanding tasks~\cite{zhao2019real,wu2024real,schmidt2023sendd}. Kurmann~\cite{kurmann2017simultaneous} used a UNet-based architecture for simultaneous tool recognition and pose estimation. Du et al.~\cite{du2018articulated} used Fully Convolutional Networks (FCN) for articulation-agnostic multi-instrument pose estimation and evaluated it on different datasets. Kayhan et al.~\cite{kayhan2021deep} proposed a deep attention-based semi-supervised method for 2D surgical instrument pose estimation. Li et al.~\cite{li2022multi} proposed a semi-supervised multi-task learning framework for joint pose estimation and segmentation. Lu et al.~\cite{9714837} explored the sim-to-real transfer for robot keypoint detection via domain randomization.

\subsection{Datasets for Surgical Tool Pose Estimation}
There are many datasets and benchmarks for the human hand~\cite{hampali2022keypoint}, whole body~\cite{andriluka2018posetrack}, and animal~\cite{yang2022apt} pose estimation. However, the public datasets for articulated surgical tool pose estimation are limited. The Retinal Microsurgery Instrument Tracking (RMIT)~\cite{sznitman2012data} dataset consists of three video clips of 480p resolution during \textit{in vivo} retinal microsurgery. A single instrument with 4 annotated joint positions is presented in this dataset. Table~\ref{table_1} provides a comparison between our SurgPose dataset versus other existing datasets that exclusively consider robotic instruments.

The Instrument Segmentation and Tracking challenge at EndoVis15~\cite{bodenstedt2018comparative} provides a dataset for articulated surgical tool tracking with only shaft point annotation.  To estimate the articulation of the instrument, Du et al.~\cite{du2018articulated} relabeled the EndoVis15 dataset by manually annotating 1850 frames with 5 keypoints. The SurgRIPE~\cite{surgripe2024xu} challenge at EndoVis23 released a dataset for 6D pose estimation of the wrist of da Vinci surgical instruments. SurgRIPE provides high-quality images captured from the da Vinci Si endoscope with 6D pose annotations obtained from a special keydot pattern. This pattern is removed by inpainting. In EndoVis24, the PhaKIR sub-challenge~\cite{phakir2024rueckert} release a new dataset of 13 cholecystectomy videos collected in three hospitals. PhaKIR includes 30K frames with multi-instrument multi-joint keypoints annotations at an interval of one frame per second, but the instruments are conventional instead of robotic. Furthermore, the monocular laparoscopes used in this dataset are uncalibrated. 

\subsection{Fluorescent Markers}
Manually labeling keypoints is time-consuming, cumbersome, and prone to error. To address this, Thananjeyan et al.~\cite{thananjeyan2022all} proposed LUV, a labeling methodology using a transparent and UV-fluorescent paint. They show LUV's feasibility for collecting labels of fabric keypoints, surgical thread, and suturing needles. In endoscopy, Schmidt et al. developed STIR~\cite{schmidt2024surgical}, a dataset of videos of deformable surgical tissue. By tattooing the tissue with Infrared (IR) fluorescent dye, indocyanine green (ICG), STIR generates markers that are invisible to visible spectrum algorithms. STIR markers can be detected under the IR spectrum using da Vinci Xi's Firefly mode.

\section{Methodology}
\begin{figure}[!h]
\centering{\includegraphics[width=0.48\textwidth]{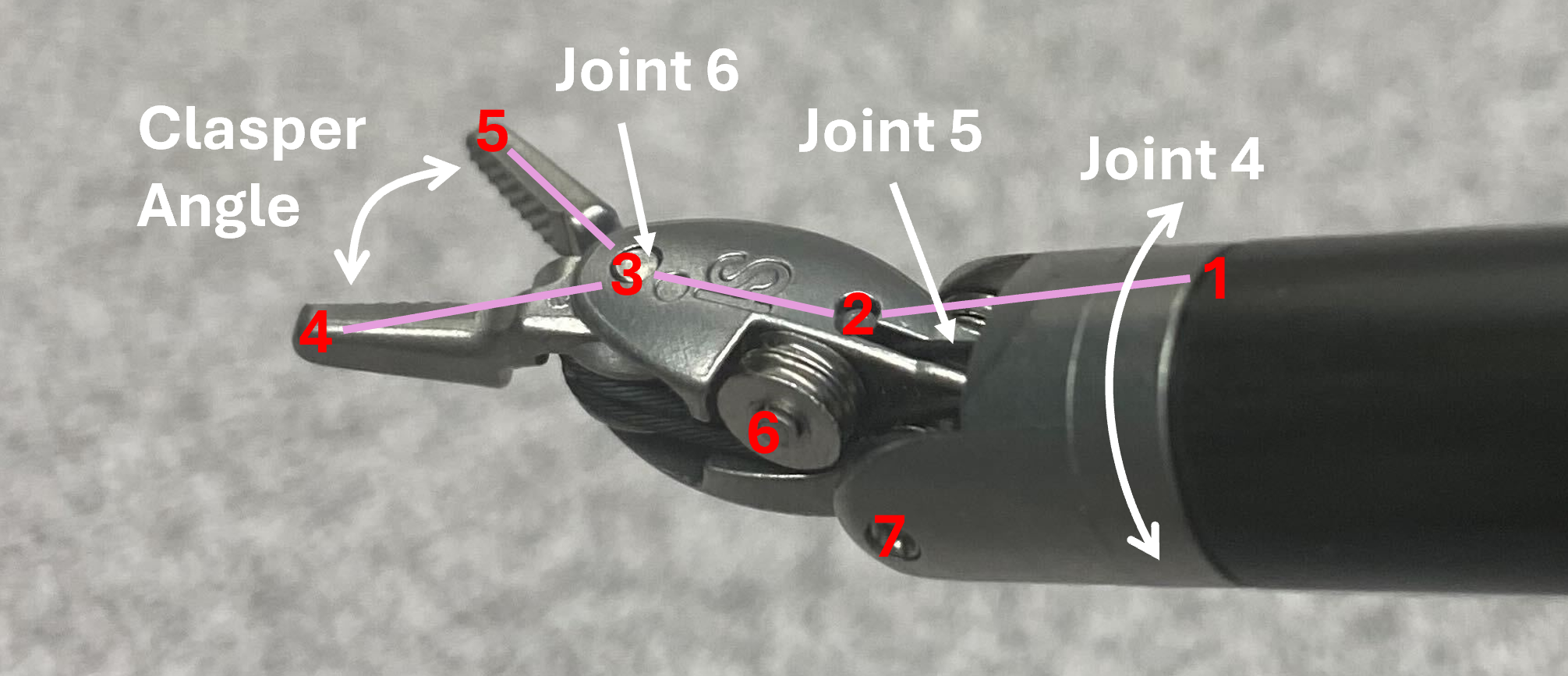}}
  \caption{Keypoint definition for the large needle driver. The instrument pose is represented with a skeleton capturing its joints and links. Similarly to the large needle driver, all instruments in SurgPose have 5 keypoint (1-5) skeletons. Keypoints 6 and 7 are redundant, but may be helpful when conducting tracking.}
  \label{fig2}
\end{figure}

In this section, we describe how we construct the SurgPose dataset and verify the visibility of the transparent UV-reactive markers. 
Fig.~\ref{fig2} shows an example of the keypoints (red font) of articulated tools in SurgPose. We first present the labeling methodology based on UV-reactive paint followed by PSM trajectory generation. We then present the experiment setup and data collection pipeline in detail. Next, we describe how we extract keypoint annotations along with other data processing methods including stereo matching. Finally, we describe how we validate the invisibility of the UV markers in conventional camera images.

\subsection{UV Reactive Paint Labeling} 
\begin{figure}[!h]
\centering{\includegraphics[width=0.48\textwidth]{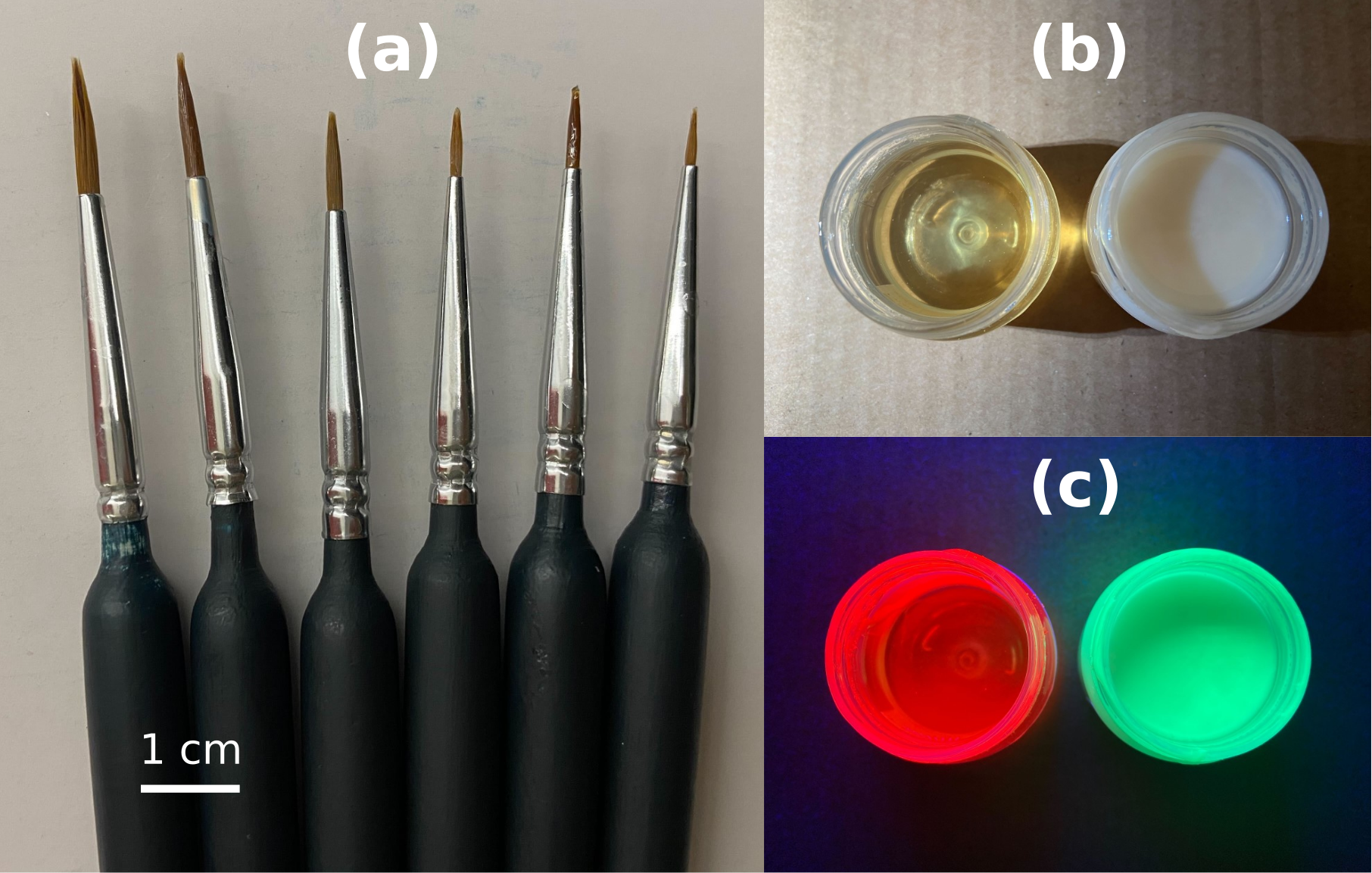}}
  \caption{(a) The miniature paint brushes are used for marking keypoints. (b) The UV reactive paint under the white light, the left and right are red and green, respectively. (c) The UV reactive paint under the black light. }
  \label{fig3}
\end{figure}
The data collection framework of SurgPose is designed in a semi-autonomous manner. We aim to reduce human involvement in labeling as much as possible while maintaining the data and annotation quality. Given this motivation, there are two criteria for the UV-reactive paint: 1) to minimize the image deviation introduced by the invisible marker, the appearance difference of the surgical instrument before and after marking should be as minimal as possible; 2) to reduce the human labeling effort, the fluorescence under black light should be apparent enough so that each keypoint can be easily extracted by simple image thresholding or an off-the-shelf segmentation method. According to the recommendation in LUV~\cite{thananjeyan2022all}, we adopt lacquer-based UV-reactive paint (GLO Effex, Murrieta, CA, USA). We tested two types of paint, red and green. The red paint has a more transparent texture while the green paint is more opaque. Since both paints have desired fluorescence, we chose the less visible red paint for SurgPose data collection. To easily paint the keypoints, we selected miniature paintbrushes (SAVITA, China). 
Technically, this lacquer-based paint can be removed by any Acetone cleaner. 

\subsection{PSM Movement Trajectory Generation}
After marking the keypoints of the surgical instrument, we use the dVRK~\cite{kazanzides2014open} to control the PSM~1 and PSM~3 moving along off-line generated trajectories.
The trajectory consists of 1) the position trajectory (without rotation) of the end-effector, which is at the center of joint 6 in Fig. \ref{fig2}, in Cartesian space; 2) the configuration space (C-space) trajectory with respect to joints 4-6, and the gripper angle, denoted as joint 7.
The position trajectory should be distributed in the workspace as evenly as possible to cover the edges of the camera view. 
In order to measure keypoints and gather visible images, the PSMs need to execute each trajectory twice, once under white and once under UV light.
To ensure repeatability, we require the trajectory curve to be closed and smooth in both Cartesian space and C-space.
To avoid collisions, we separate the workspace of PSM~1 and PSM~3 without overlap. 
Since these two PSMs are left and right symmetric, we will use a single PSM as an example to introduce the random trajectory generation strategy.

To generate the position trajectory of the end-effector, we first define a rectangular workspace for the PSM by manually moving the PSM to its vertices. 
Then we randomly select trajectory points in this workspace and use a periodic interpolating cubic spline~\cite{lee1989choosing} to smoothly connect these random points. Afterward, we interpolate this curve with a proper step size (1000 steps in this study) to generate a smooth, closed 3D position trajectory.  For joints 4-7 we generate periodic sinusoidal trajectories. For each step, we first change the joint states and then move the end-effector to the target position. The robot's motion is obtained as follows:
\begin{algorithmic}
\State $Step \gets 0$, 
\While{$Step \leq 1000$} 
    \State 1) Change joint states
    \State 2) Move end-effector position
    \State 3) Measure and record joint states of 4-7 and end-effector pose (6D pose $\in$ SE(3))
    \State 4) Capture left and right frames
\EndWhile
\end{algorithmic}


\subsection{Experimental Setup}
\begin{figure}[!h]
\centering{\includegraphics[width=0.48\textwidth]{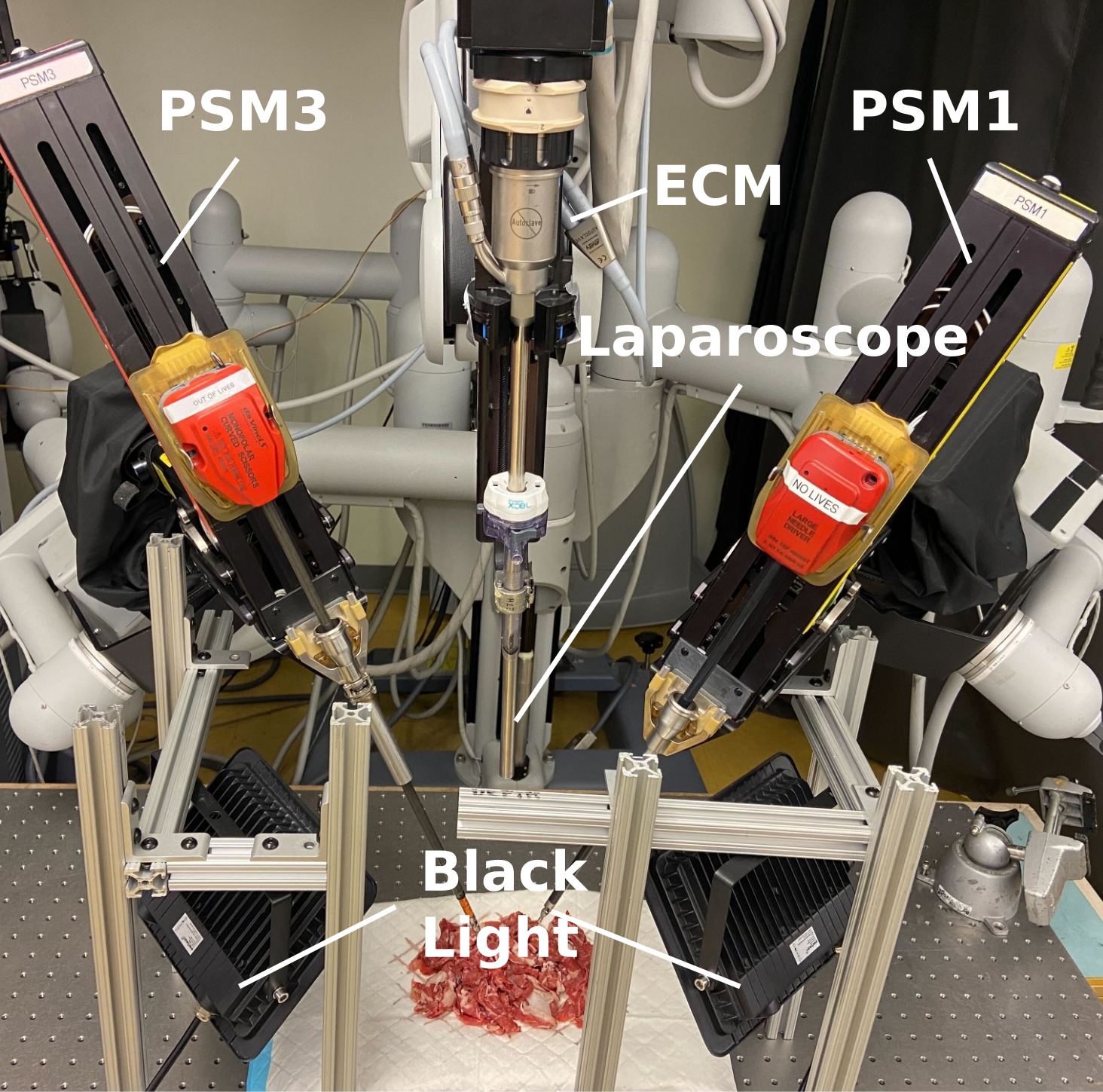}}
  \caption{Experimental setup. PSM~1 and 3 are controlled by the dVRK. The ECM is always static during the data collection. When the surgical lamp is turned on, black lights are turned off, and vice versa.}
  \label{fig4}
\end{figure}

As depicted in Fig. \ref{fig4}, the experimental setup consists of a da Vinci IS1200 system (first generation) with the dVRK and two 365nm 50W UV LED black lights (Everbeam, Surrey, BC, Canada). The black lights are attached to Aluminum rail frames that can adjust the lamp pose flexibly to make sure that the UV-reactive markers are illuminated properly.

\subsection{Data Processing and Annotation}
The goal of annotation is to extract the semantic keypoints of the surgical instrument from the video collected under UV black light. A simple method is to segment fluorescent markers and take their centers as keypoints. This segmentation can be achieved by doing a simple thresholding for each frame. However, we cannot directly obtain the semantic label of the keypoints in this way. Furthermore, the marker's fluorescence intensity varies in different poses and videos. Therefore, we use the off-the-shelf foundation model Segment Anything~2 (SAMv2)~\cite{ravi2024sam}. Similarly to other semi-supervised video object segmentation methods, we need to provide prompts for each marker dot in the first frame of the video sequence; then, SAMv2 can propagate the mask over the entire video. Through multiple experiments, we determined that SAMv2 can segment very weak fluorescent markers. This property of SAMv2 ensures labeling efficiency. 
We reviewed and corrected every labeled frame from SAMv2's label propagation. 
We further leverage SAMv2 on the visible video sequences to segment the instruments. 
This provides us with the area of the projected instrument in each frame, and can use this area as a scaling parameter when we determine keypoint detection metrics. 
Indeed, keypoint localization should scale with object size.
Further segmentation (in part-level) of the surgical instrument individual links can provide insight into designing novel metrics, customized for pose estimation of each link of the articulated tool. 

\subsection{Camera Rectification and Stereo Matching}
In SurgPose, we calibrate the stereo laparoscope using a checkerboard and the OpenCV library. Given the intrinsic parameters of the stereo cameras and the rigid transformation between them, we perform image rectification. Given the rectified stereo, we use the RAFT~\cite{teed2020raft} method to estimate the disparity between the left and right images. Finally, we convert the disparity to depth by the following formulation:
$$
Depth=\frac{focal\_length\times baseline}{\left|disparity\;+\;(c_{x1}-c_{x0})\right|}
$$
in which the $c_{x1}-c_{x0}$ is the difference of principal points along the $x$ axis.

\subsection{Quantifying the Visibility of Markers}
We carefully reviewed every frame in SurgPose.
The majority of markers were invisible in the laparoscopic video; nevertheless, in some instances of overpainting, the keypoint markers could be seen with the naked eye. 
To explore the visibility of the UV fluorescent markers, we used both quantitative and qualitative evaluations. 
  
First, we obtained images of the surgical tool with and without the UV fluorescent markers for comparison. We attached the instrument to the robot and captured the laparoscopic frame without any UV markers. Then, we used the brush to paint the keypoints and captured the frame after painting. Both frames are collected in white light. 
For each type of instrument, we recorded these pairs of images under 100\% and 50\% lighting. There are 12 frame pairs in total. 
We subtract each pair of frames to get the pixel difference. By observing this difference, we can qualitatively determine whether the markers are visible.
Since the instruments and lighting conditions can change slightly between the different acquisition times of the laparoscopic images, it is not sufficient to just characterize the intensity of the pixel difference images. 
Similar to STIR~\cite{schmidt2024surgical}, we calculate the Gray Level Co-occurrence Matrix (GLCM) features~\cite{haralick1973textural} to assess the visibility of markers. 
GLCM features can represent fine-grained texture pattern. This property makes GLCM suitable for medical image analysis, in which the features are often subtle. 
We crop the image patch surrounding each keypoint and calculate the GLCM features for each patch. We then compute the dissimilarity and correlation of the GCL matrices.
We can then train a classifier to distinguish the image patches in the GLCM feature space, to see if visibility can be estimated algorithmically.   

\subsection{Dataset Details}
Table \ref{table_2} summarizes the instance number of each instrument category in the training and validation set. SurgPose is presented as a set of folders associated with a trajectory. 
The keypoint annotation follows the COCO format~\cite{lin2014microsoft}. 
Each trajectory includes 1001 steps associated with the left and right images along with the end-effector forward kinematics and joint states of PSM~1 and 3. 
The execution time of each trajectory is approximately 7 minutes.
The resolution of the images is 986$\times$1400. 
The training set backgrounds include chicken gizzard and pork liver \textit{ex vivo}, while the validation set includes chicken thighs and beef \textit{ex vivo} background. The surgical lamp intensity is 100\% in the training set and 50\% in the validation set. When using 2D pose estimation and tracking methods, one can choose to use the left frames. We provide a stereo-matching baseline based on RAFT as a reference depth. 
We note that there are many other depth estimation algorithms that could be used to estimate depth for 3D tasks from the stereo pairs provided in our dataset. 
\begin{table}[h]
\caption{Instance Statistics of Different Instruments}
\label{table_2}
\begin{center}
\begin{tabular}{l c c}
\toprule
Tool Categories & Training Instances & Testing Instances   \\
\midrule
Large Needle Driver & 32K & 12K\\
Mega Needle Driver & 12K & 4K  \\
Micro Forceps & 6K & 8K \\
Curved Scissor &  6K & 4K\\
DeBakey Forceps & 12K & 4K \\
Prograsp Forceps & 12K & 8K \\
\bottomrule
\end{tabular}
\end{center}
\end{table}

\section{EXPERIMENTS AND RESULTS}
In this section, we discuss experiments and results from three perspectives to verify the usability of SurgPose: 1) the invisibility of UV-reactive markers; 2) the effectiveness of data collection and keypoints annotation extraction; and 3) baseline results on the SurgPose dataset.

\subsection{Invisibility of Markers}
\begin{figure}[h]
\centering{\includegraphics[width=0.48\textwidth]{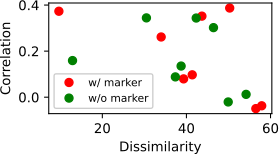}}
  \caption{The GLCM features of the image patches around the keypoints.}
  \label{fig5}
\end{figure}

\begin{figure*}[h]
\centering{\includegraphics[width=\textwidth]{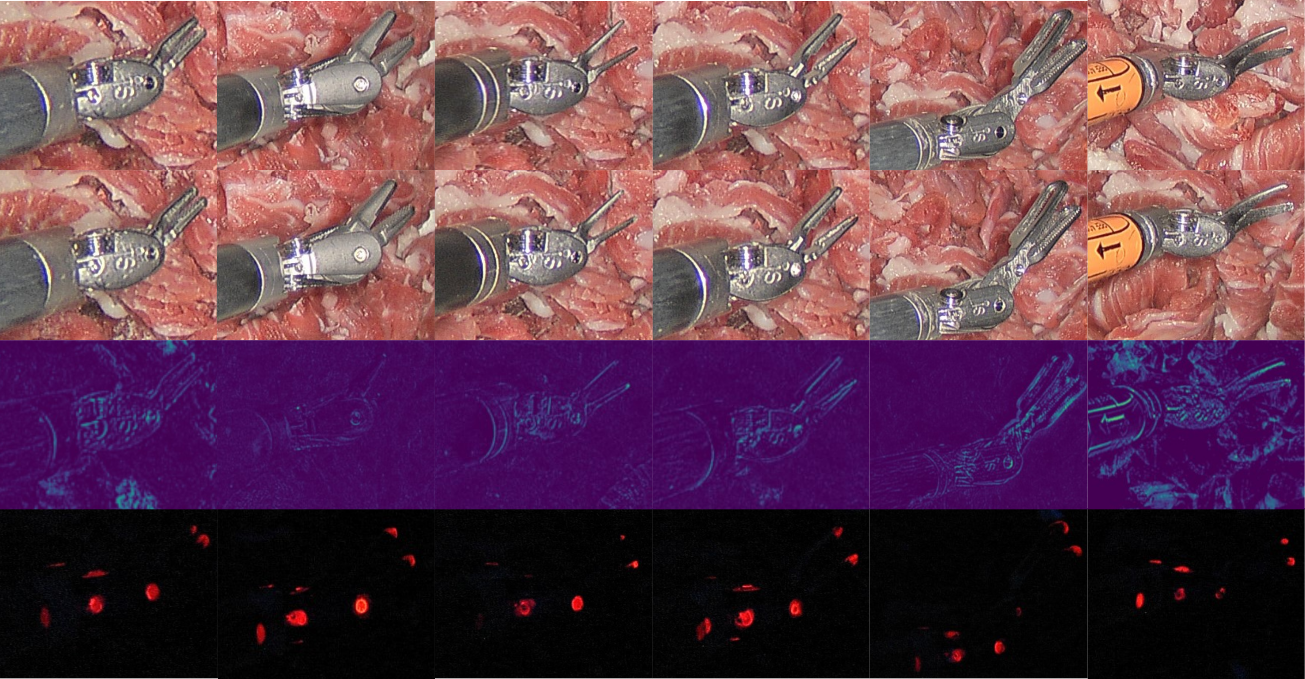}}
  \caption{Comparison of the tool appearance before and after marking by UV reactive paint. The top row to the bottom are frames before marking, frames after marking, the difference between them, and the frames under the UV light, respectively.}
  \label{fig6}
\end{figure*}

As shown in the third row of Fig. \ref{fig6}, there is no significant difference in the locations painted by the UV fluorescent marker. The difference in the edge region of the instrument is mainly from the motion caused by the painting. 
To calculate the GLCM features, we sample 40$\times$40-sized image patches around the keypoints and convert them to grayscale. 
Dissimilarity and correlation features of the GLCM matrices are used to represent the image patches. The results are shown in Fig. \ref{fig5}. Note that we only show the GLCM features of the first column image patches in Fig. \ref{fig6} as an example. 
The image patches are intertwined in the GLCM feature space and are difficult to divide using Linear/Quadratic Discriminant Analysis or clustering algorithms.

\subsection{Feasibility of Data Collection}
Despite the powerful performance of SAMv2 on video objection segmentation, human involvement (correction and intra-video prompt) is still necessary to ensure the accuracy of the keypoint annotations. We record the number of human involvements, minus those for the initial frame, to analyze the human labor for labeling. 
Human correction was needed on average 2.1$\pm$1.6 times for each video, requiring 1 to 10 minutes annotation time.  

\subsection{Analysis of Baselines}
We evaluate the pose estimation performance of three baselines, YOLOv8~\cite{varghese2024yolov8}, ViTPose~\cite{xu2022vitpose}, and DeepLabCut~\cite{mathis2018deeplabcut} on SurgPose. 
For YOLOv8, we adopt the YOLOv8-pose-x architecture.
For ViTPose, we use pre-trained MAE weights corresponding to ViT-Pose-H. For DeepLabCut, we use weight initialization from ImageNet and the ResNet-101 architecture for the detector.

Since these baseline methods are originally designed for human pose estimation and are pre-trained on human pose datasets, we adopt the standard metric Object Keypoint Similarity (OKS) defined by COCO~\cite{consortium2014eval}. OKS can be calculated using the Euclidean distance between a predicted and a ground truth point, which is passed through an unnormalized Gaussian distribution where the standard deviation corresponds to the square root of the size of the segmentation area multiplied by a per-keypoint constant. 
Similarly to assessing object detection tasks, thresholding the OKS defines matches between the ground truth and estimated keypoints and allows computing precision-recall curves. We adopt the COCO primary metric $mAP$, which is the mean average precision (AP) over multiple OKS thresholds (0.5:0.05:0.95). The results are shown in Table \ref{table_3}. We also show the runtime of each architecture. YOLOv8 and ViTPose can run pose estimation on SurgPose in a real-time manner. The slow inference speed of DeepLabCut is due to the full-size input image.

\begin{table}[h]
\caption{Baseline Results and Runtime on SurgPose}
\label{table_3}
\begin{center}
\begin{tabular}{c c c}
\toprule
Method & $mAP$ & Inference Latency (ms)  \\
\midrule
YOLOv8~\cite{varghese2024yolov8} & 63.3$\pm$1.23 & 16.4 \\
ViTPose~\cite{xu2022vitpose} & 53.6$\pm$1.06 & 7.7 \\
DeepLabCut~\cite{mathis2018deeplabcut}& 66.2$\pm$2.01 & 194.8 \\

\bottomrule
\end{tabular}
\end{center}
\end{table}

   
\section{LIMITATIONS AND FUTURE WORK}
Even though the size of the dataset is much larger than the existing datasets for surgical tool pose estimation, the data distribution is still not diverse enough. 
This makes deep learning models prone to overfitting. 
Baseline results demonstrate serious overfitting when tested on videos with different \textit{ex vivo} backgrounds.
This means SurgPose on its own is probably not able to train a model that can be transferred to real clinical applications. In future work, we will include additional data with more types of instruments and \textit{in vivo} cadaver environments for a more comprehensive evaluation. As well, we will include instrument trajectories that also occlude each other. It may be possible to also add specific surgical tasks, such as suturing, by using time-multiplexed lighting.
Nevertheless, SurgPose is still a valuable dataset for the design and evaluation of few-shot learning, unsupervised learning, or fine-tuning Foundation Models. Future work may introduce synthetic data from simulation software or generative AI models to develop a unified model. Diffusion model-based data augmentation is a promising technique. Its application in the surgical scene understanding is under exploration. Developing multi-modal methods to leverage the knowledge of kinematics and joint states is another task worth further exploration. 

\section{CONCLUSIONS}
In summary, we created a new dataset SurgPose that enables researchers to better develop and evaluate methods for articulated surgical tool pose estimation.


\addtolength{\textheight}{0cm} 



\bibliographystyle{IEEEtran}
\bibliography{References}

\end{document}